# THE ROLE OF RADIOMETRIC FEATURES IN CROSS-SITE LEAF-WOOD SEGMENTATION OF LIDAR POINT CLOUDS

Roman Kaharlytskyi
Dept. of Applied Mathematics, Faculty of Mathematics
University of Waterloo
Waterloo, Canada
rkaharlytskyi@uwaterloo.ca

Derek T. Robinson
Dept. of Geography and Environmental Management, Faculty of Environment
University of Waterloo
Waterloo, Canada
dtrobins@uwaterloo.ca

Roberto Guglielmi
Dept. of Applied Mathematics, Faculty of Mathematics
University of Waterloo
Waterloo, Canada
roberto.guglielmi@uwaterloo.ca

***Abstract*—Leaf-wood segmentation of individual trees from LiDAR point clouds is essential for quantitative structure models (QSMs) used in non-destructive biomass estimation. Existing segmentation methods typically exclude radiometric features (e.g., intensity, return number) to maximize cross-sensor compatibility. We challenge this design choice by evaluating cross-site and cross-platform generalization: training on the public Heidelberg dataset (terrestrial TLS, 1550nm) and testing on a novel dataset from Ontario, Canada (RPA-LS, 905nm). Results show that geometry-only methods - including state-of-the-art deep learning models trained on high-density LiDAR datasets - fail to generalize to the sparse, top-down geometry of aerial scans, achieving F1 scores ≤ 0.56. Incorporating radiometric features (intensity, return number, number of returns) improves F1 to 0.61, but more critically, increases wood recall by 119% from 0.16 to 0.35. Furthermore, geometry-only approaches often result in fragmented stem and branch components. We find that leveraging radiometric features preserves greater structural connectivity, resulting in more coherent architectures that are better suited for QSM reconstruction. We demonstrate that while geometric patterns are view-dependent and prone to overfitting scan patterns, radiometric features encode physical material properties that generalize across disparate sensors and environments.**

## I. INTRODUCTION

Tree biomass estimation is fundamental to forest inventory and carbon accounting, with Quantitative Structure Models (QSMs) serving as a primary non-destructive approach [1]. QSMs reconstruct tree topology by fitting cylindrical primitives to LiDAR point clouds; however, their accuracy depends on a "wood-only" input to prevent foliage from distorting geometric reconstruction. Consequently, the semantic segmentation of leaf and wood points is a critical prerequisite. While this task has been addressed using classical geometric [2,3], machine learning [4], and deep learning methods [5,6], producing segmented outputs with the structural integrity required for QSM modeling remains a technical challenge.

Approaches to leaf-wood segmentation differ primarily in their feature representations. Geometry-based methods rely solely on spatial coordinates, deriving local descriptors such as linearity, planarity, and surface variation [2,3]. Radiometric methods instead exploit intensity-based reflectivity differences between wood and foliage [7]. Hybrid methods combine both representations [4,8].

Recent state-of-the-art deep learning methods have diverged in their approach to radiometric features. ForestFormer3D [6] explicitly uses only XYZ coordinates, arguing that intensity and return attributes "are not consistently available" and excluding them ensures "broad applicability." However, these models are typically trained on dense, ground-based terrestrial laser scanning (TLS) datasets. When applied to Unmanned Laser Scanning (ULS) data - which exhibits different occlusion patterns, lower point densities, and top-down viewing angles - their learned geometric representations often fail to transfer. Similarly, other software like PointsToWood [5] incorporate reflectance through a gated module but defaults to coordinate-only processing when reflectance is unavailable or inconsistent.

We challenge the assumption that geometric features alone are sufficient for robust leaf-wood segmentation. Radiometric features encode physical material properties: wood exhibits higher intensity than foliage (in NIR/SWIR wavelengths), a relationship that persists regardless of viewing angle. While geometric descriptors (e.g., verticality, sphericity) are highly sensitive to scan pattern (TLS concentric rings vs. ALS linear scan lines), radiometric responses represent fundamental biological differences. Therefore, radiometric features may offer superior cross-platform generalization despite sensor variability.

We investigate cross-platform transferability through a transfer learning experiment: training on European TLS data (1550nm) and testing on Canadian RPA-LS data (905nm). We compare geometry-only versus radiometrically-augmented models, quantify performance gains from radiometric features including intensity, return number, and number of returns, and evaluate whether standard metrics capture the structural connectivity required for QSM reconstruction.

## II. MATERIALS AND METHODS

### A. Datasets

Training Data: The publicly available TLS leaf-wood dataset from Weiser et al. [9] comprises 11 trees of 7 European species scanned with a RIEGL VZ-400 terrestrial laser scanner, containing approximately 50 million manually labeled points.

Test Data: We introduce 17 labeled trees from Hullett Provincial Wildlife Area, Southern Ontario, Canada (see Fig. 1 for study area map), acquired via RIEGL Ultra 120 mounted on a Harris Aerial Carrier H6 Remotely Piloted Aircraft (RPA) platform. The mixed deciduous forest at Hullet includes Sugar Maple, Green Ash, American Elm, and American Basswood. Point density acquisition fell within 10,000–12,000 pts/m². Ground truth labels were generated via manual segmentation.

This configuration represents cross-site transfer learning: European mixed forest (TLS) → North American hardwoods (RPA-LS), with different scanner models and acquisition geometries.

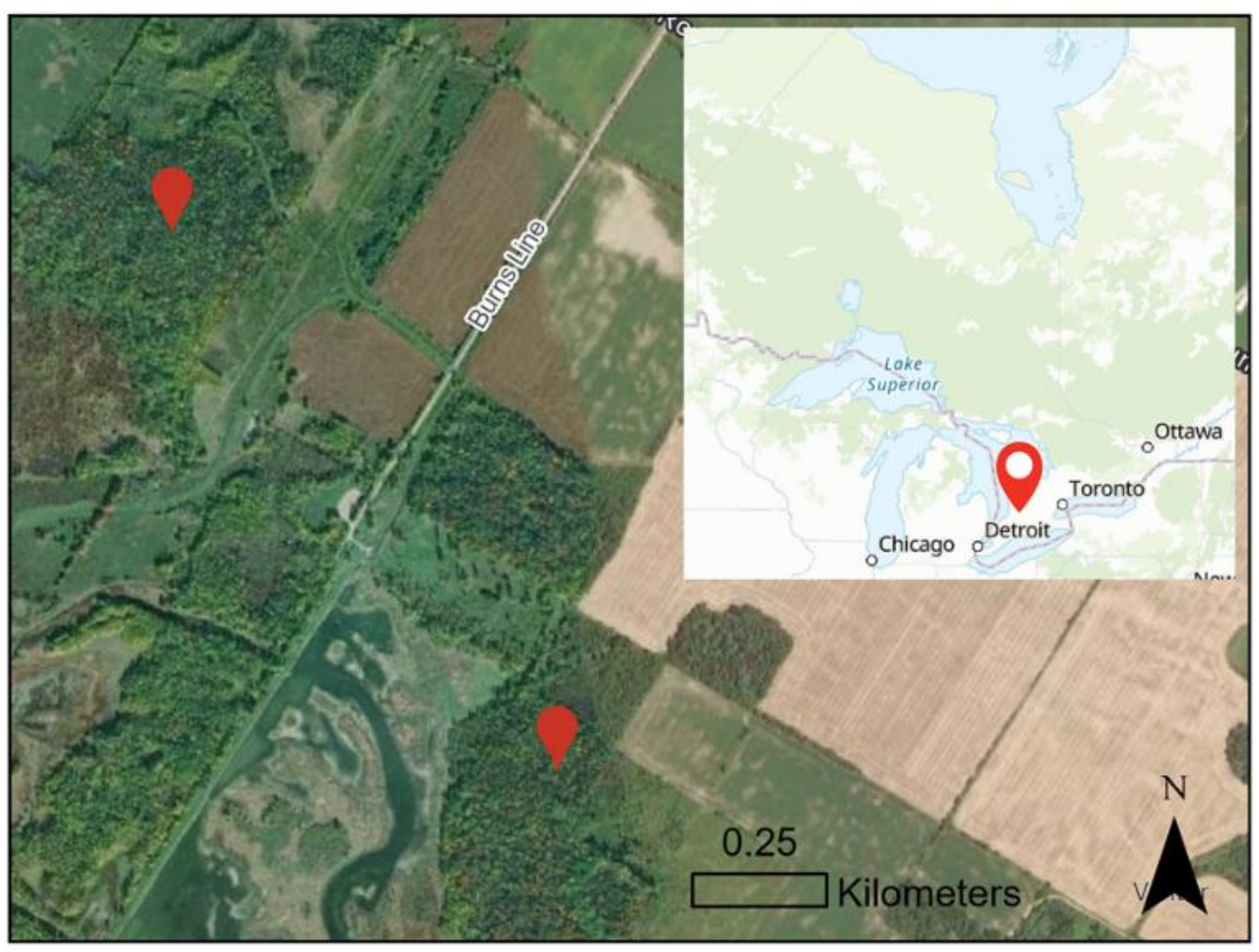

Fig. 1. Study Area comprising two forest plots in Hullet Wildlife Preserve, southwestern Ontario, Canada.

### B. Methods Evaluated

To assess whether radiometric features improve cross-platform generalization, we compared five approaches representing different paradigms: classical geometric methods, state-of-the-art deep learning models, and gradient boosting classifiers with and without radiometric augmentation:

LeWoS [2]: An unsupervised approach utilizing geometric separation based on surface variation. The segmentation threshold was optimized to 0.16 for this dataset.

PointsToWood [5]: Deep learning architecture using PointNet++ with a gated reflectance integration module. Due to domain shift from TLS to RPA-LS, the pre-trained model exhibited low confidence, producing noisy predictions at default inference settings. Through systematic threshold tuning of the implementation's confidence parameters, an any-wood probability threshold of 0.95 was found to suppress false-positive noise in the canopy while retaining high-confidence branch structures.

ForestFormer3D [6]: Transformer-based architecture using only XYZ coordinates. The pre-trained model was applied directly. Attempts to fine-tune or retrain with radiometric inputs failed to converge on the 11 training trees from Weiser et al. [9], suggesting deep learning methods require substantially larger training datasets for this domain.

CatBoost1 (Geometric): Gradient boosting classifier using 26 geometric features computed at multiple radii (0.1–0.6m): eigenvalue descriptors (linearity, planarity, sphericity, anisotropy, eigenentropy, surface variation, verticality), PCA components, neighbor counts, and relative height [4]. The training and inference pipeline is shown in Fig. 2. Data augmentation techniques were applied during training, including variable density gradients to simulate non-uniform sampling and leaf occlusion simulation to mimic canopy blockage. All features were standardized using Z-score normalization (zero mean, unit variance) prior to training and inference.

CatBoost2 (Geometric + Radiometric): Identical pipeline, augmentation strategy, and geometric features as CatBoost1, with the addition of three radiometric attributes: normalized intensity, return number, and number of returns.

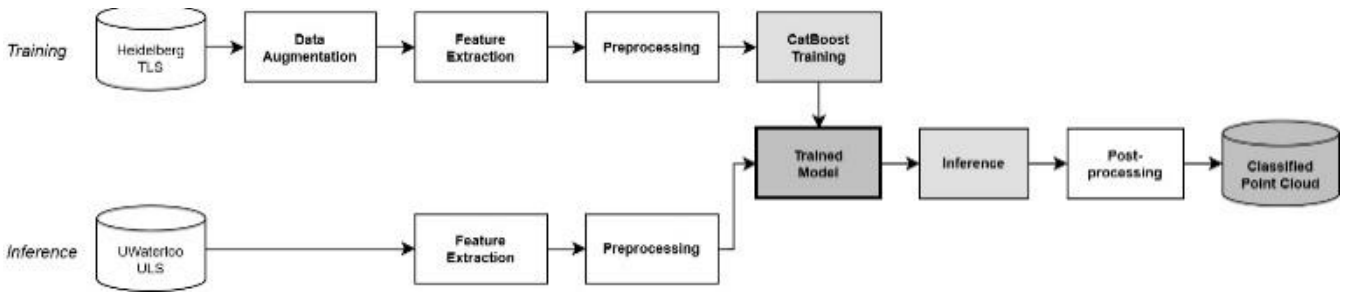

Fig. 2. CatBoost training and inference pipeline

### C. Postprocessing and Evaluation

Raw classification outputs contain fragmentation artifacts (such as disconnected branches, isolated points) that vary by method and degrade structural connectivity for QSM reconstruction. To enable fair performance comparison, all methods received identical postprocessing: neighborhood smoothing, morphological operations, connectivity analysis, and gap filling. All quantitative results and visualizations reflect post-processed outputs.

Quantitative performance was evaluated using F1 score, Jaccard Index, overall accuracy, and class-specific precision and recall aggregated across all test trees. To isolate error modes associated with canopy occlusion and varying point density, F1 scores were further stratified by vertical position. Each tree was divided into lower, middle, and upper thirds based on total tree height, separating the trunk-dominated lower and middle strata from the foliage-dense upper canopy.

## III. RESULTS

### A. Quantitative Comparison

All geometry-only methods exhibited poor cross-domain generalization, performing significantly below benchmarks reported in their respective single-domain studies (Table I). CatBoost2, which incorporates radiometric features, achieved the highest F1 score (0.606) and substantially improved Wood Recall compared to its geometry-only counterpart (0.346 vs. 0.159).

TABLE I. LEAF-WOOD SEGMENTATION RESULTS

| Method | Features | F1 | Jaccard Index | Accuracy | Wood Precision | Wood Recall | Leaf Precision | Leaf Recall |
|---|---|---|---|---|---|---|---|---|
| PointsToWood | XYZ (refl=0) | 0.352 | 0.214 | 0.354 | 0.231 | **0.479** | 0.565 | 0.298 |
| ForestFormer3D | XYZ | 0.538 | 0.424 | 0.721 | 0.717 | 0.149 | 0.722 | 0.974 |
| LeWoS | XYZ | 0.557 | 0.437 | 0.724 | 0.715 | 0.179 | 0.724 | 0.968 |
| CatBoost1 | Geometric | 0.559 | 0.444 | **0.742** | **0.986** | 0.159 | 0.729 | **0.999** |
| CatBoost2 | Geometric + Radiometric | **0.606** | **0.462** | 0.700 | 0.514 | 0.346 | **0.748** | 0.856 |

However, quantitative metrics alone do not fully capture structural differences. Visual inspection (Fig. 3) reveals two distinct failure mechanisms in geometry-only methods. PointsToWold (Fig. 3b) exhibits high commission error (false positives), misclassifying foliage as wood despite threshold tuning. This results in noisy predictions with low Wood Precision (0.231). Conversely, geometry-only machine learning methods - LeWoS, ForestFormer3D, and CatBoost1 (Fig. 3c-e) - exhibit severe omission error (false negatives). These approaches successfully isolate the main stem but systematically misclassify complex branching structures as foliage, yielding inflated Leaf Recall (>0.96) at the expense of Wood Recall (<0.18).

The inclusion of radiometric features (Fig. 3f) mitigates both failure modes, recovering fine branching architecture while reducing canopy noise. Although CatBoost2's Wood Precision (0.514) remains moderate due to some residual leaf-to-wood misclassification, the substantial gain in Wood Recall (119% increase) and preservation of structural connectivity represent a favorable trade-off for QSM reconstruction, where branch completeness is critical.

Augmenting geometric features with radiometric data (CatBoost1 → CatBoost2) improves F1 from 0.559 to 0.606 (+8.4%) and more than doubles wood recall from 0.159 to 0.346 (+117%). Fig. 4 and Fig. 5 show this improvement is consistent across all 17 test trees.

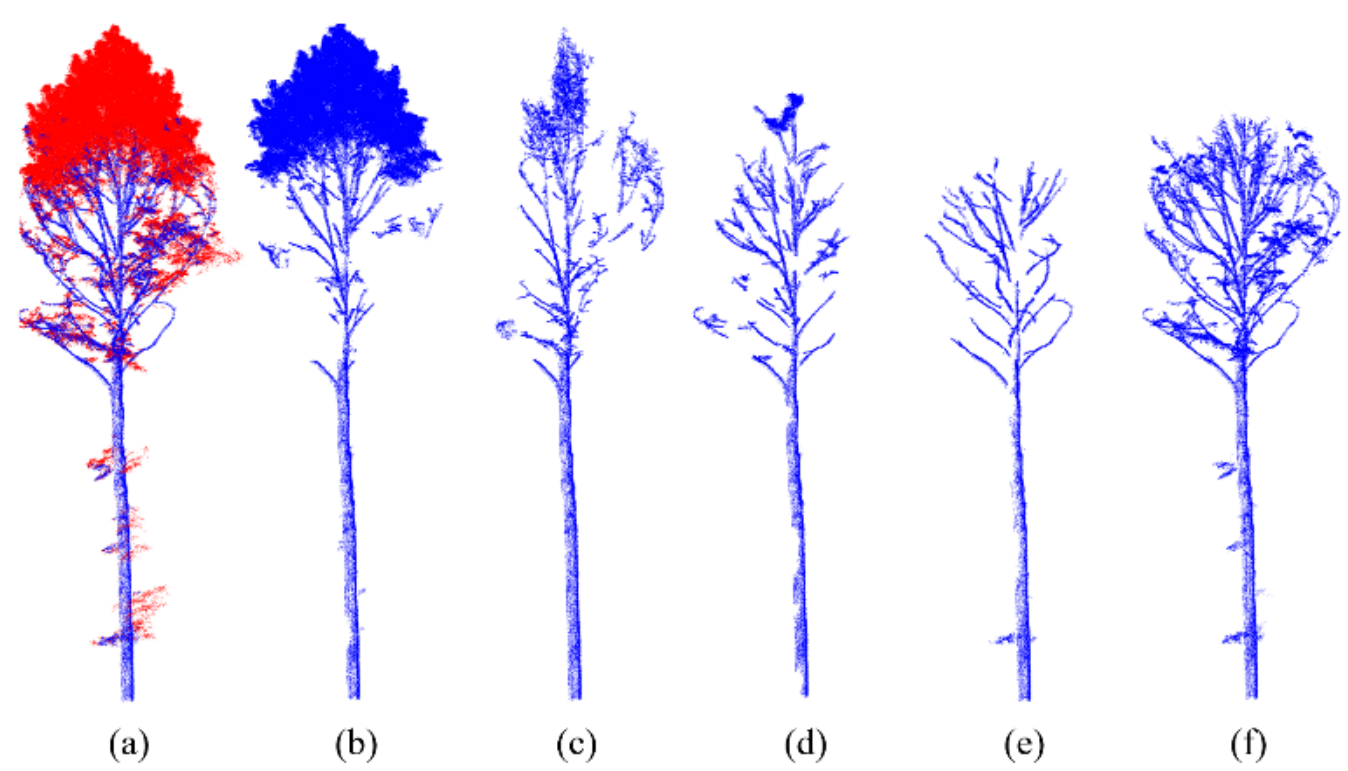

Fig. 3. Visualization of **(a)** Ground truth reference (Red: Leaves, Blue: Wood) and outputs from the five evaluated approaches **(b)-(f)** for a single tree. Predicted wood points (Blue) for each method: **(b)** PointsToWood, **(c)** ForestFormer3D, **(d)** LeWoS, **(e)** CatBoost1 (Geometric features), and **(f)** CatBoost2 (Geometric + Radiometric features).

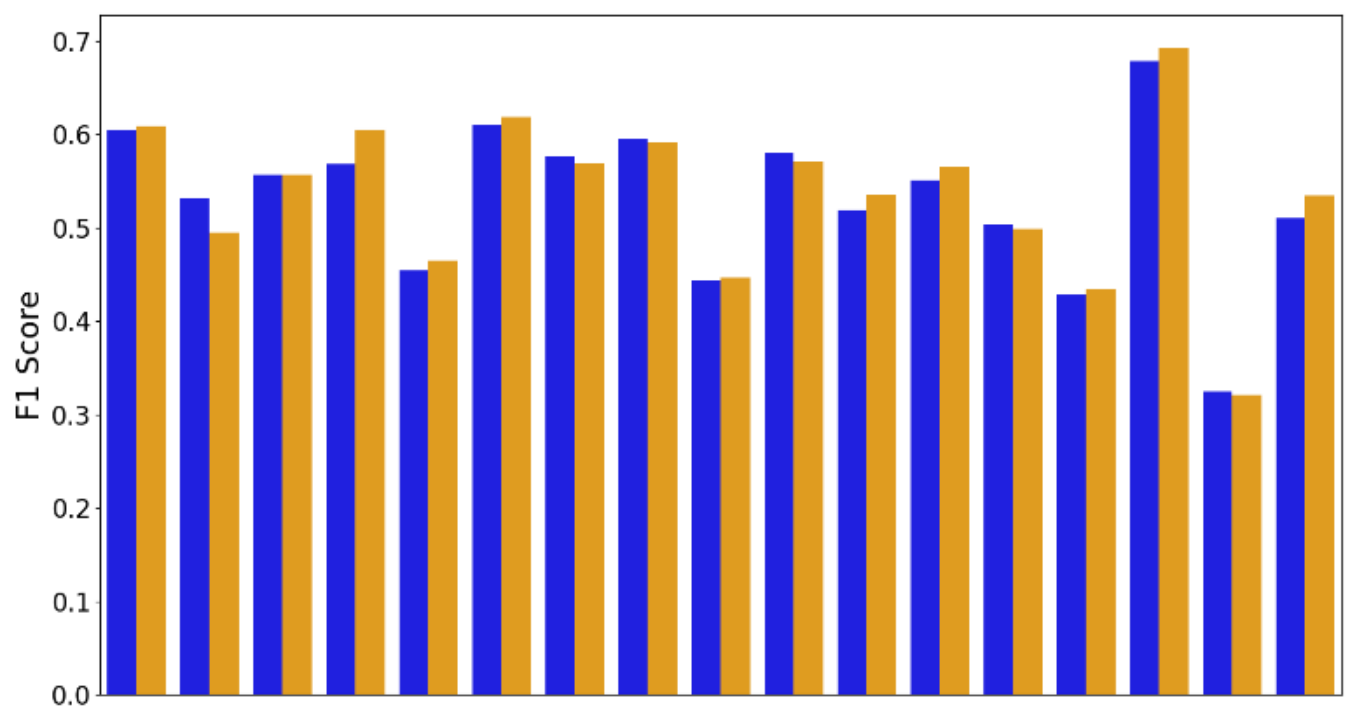

Fig. 4. F1 comparison for 17 sample trees. Blue: CatBoost1 (Geometric features), and Orange: CatBoost2 (Geometric + Radiometric features).

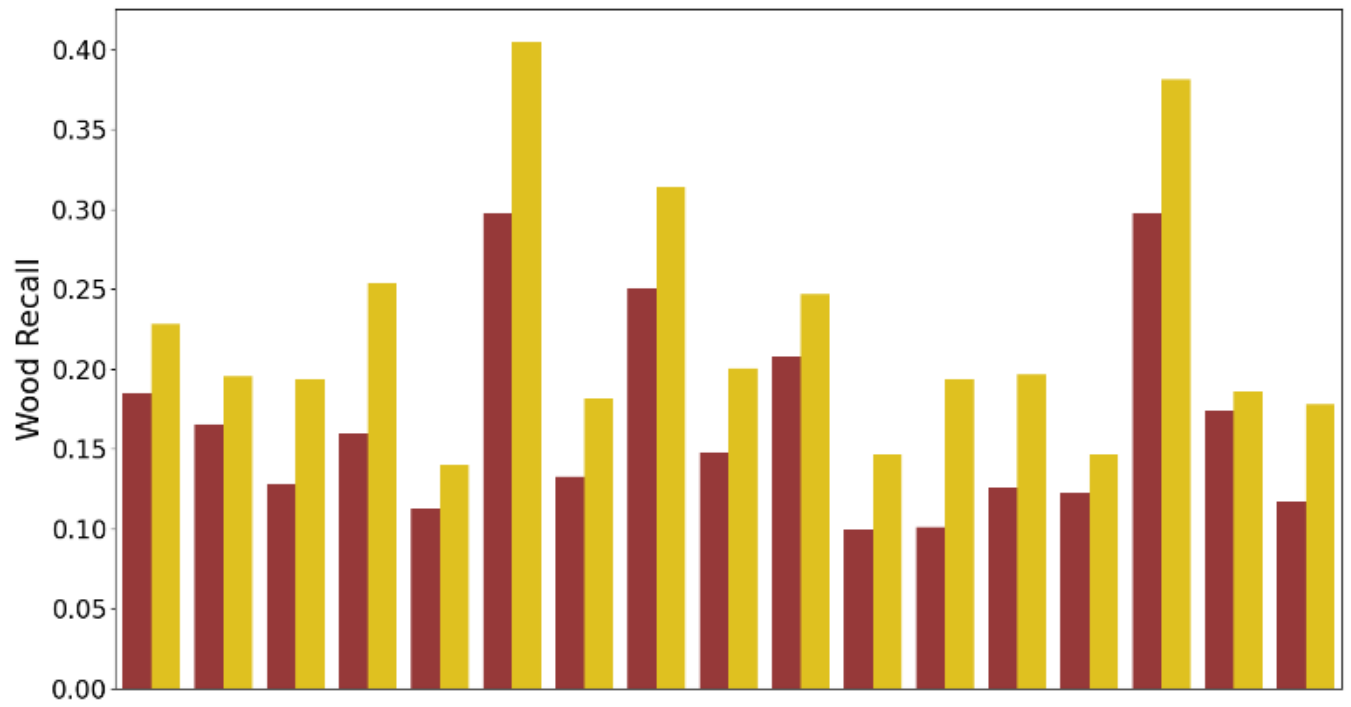

Fig. 5. Wood Recall Comparison for 17 sample trees. Red: CatBoost1 (Geometric features), and Gold: CatBoost2 (Geometric + Radiometric features).

### B. *Canopy Height Stratification*

To isolate error modes associated with varying occlusion and point density, we stratified F1 scores by relative height within the tree crown (Table II). Each tree was divided into lower, middle, and upper thirds based on total height.

Geometry-only methods show declining performance in the upper canopy (F1 ~ 0.46-0.55), where woody structures are heavily occluded by foliage, resulting in sparse, disconnected returns. In contrast, the radiometrically-augmented model (CatBoost2) achieves substantially higher performance in the upper canopy (F1 = 0.708), suggesting robustness to sparse geometric context.

TABLE II. F1 SCORES BY CANOPY HEIGHT STRATUM

| Method | Lower | Middle | Upper |
|---|---|---|---|
| PointsToWood | 0.710 | 0.573 | 0.458 |
| ForestFormer3D | 0.534 | 0.540 | 0.464 |
| LeWoS | 0.473 | 0.581 | 0.514 |
| CatBoost (Geometric) | **0.731** | **0.624** | 0.547 |
| CatBoost (Geometric + Radiometric) | 0.537 | 0.568 | **0.708** |

### C. *Feature Importance*

Feature importance analysis reveals substantial differences between geometric and radiometric models (Fig. 6). In the geometry-only model, surface_variation (19.6%), verticality (17.9%), and relative_height (14.0%) dominate. When radiometric features are included, intensity becomes the second-most important feature (18.9%), surpassed only by surface_variation (19.5%). Notably, verticality importance drops from 17.9% to 10.0%. Return attributes (number_of_returns: 1.3%, return_number: 0.6%) contribute modestly.

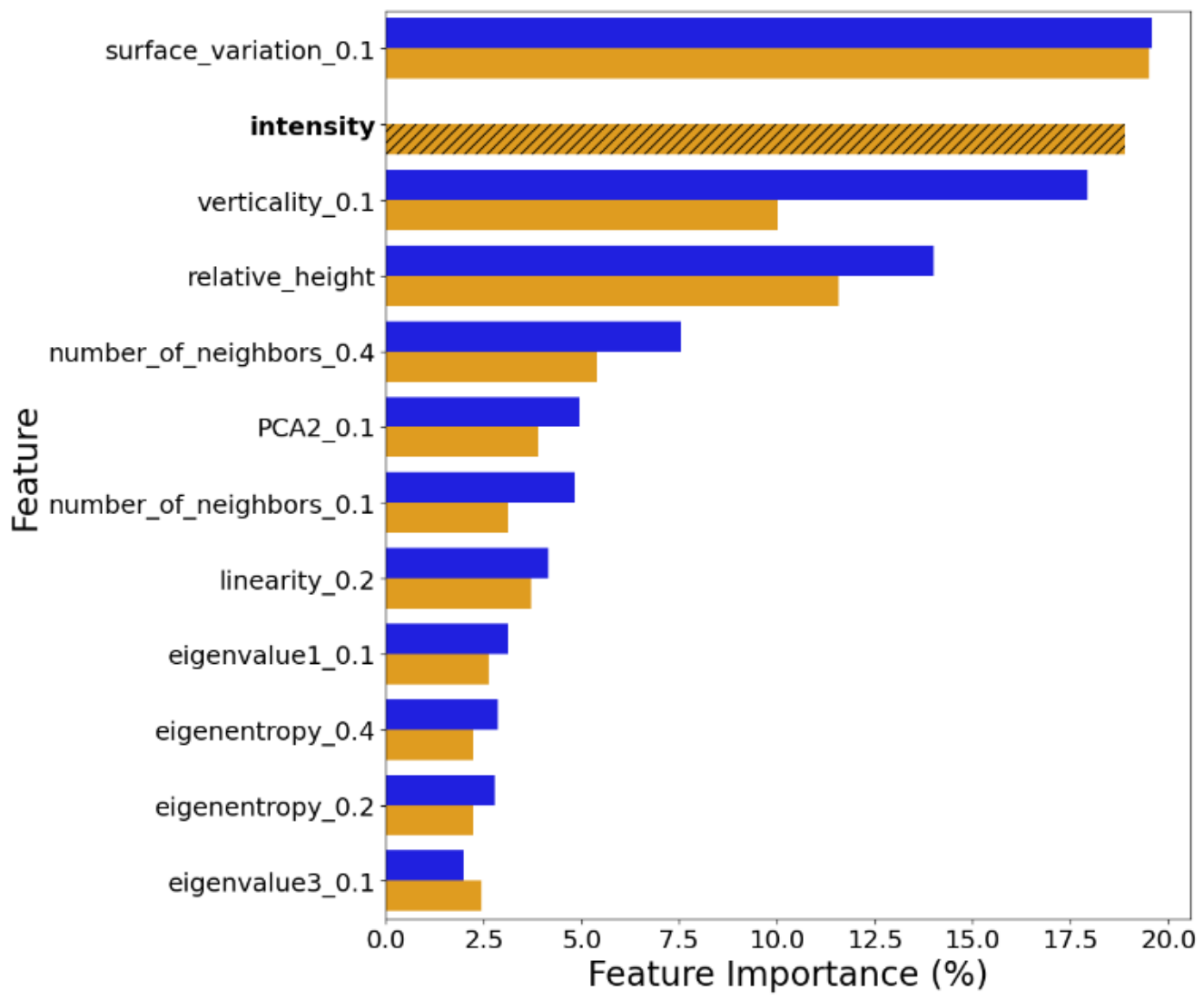

Fig. 6. Feature Importance Comparison. Blue: CatBoost1 (Geometric features), and Orange: CatBoost2 (Geometric + Radiometric features).

## IV. DISCUSSION

The limited cross-platform transferability of geometry-only methods highlights a fundamental limitation in sensor-agnostic design: geometric overfitting to scan patterns.

Geometric features are inherently view-dependent. Descriptors like verticality, sphericity, and point density distribution are functions of scanner position relative to the tree. Models trained on TLS data learn upward-looking patterns. When applied to RPA-LS data with top-down viewing geometry, these signatures are absent. This structural domain shift likely explains why ForestFormer3D and CatBoost1 failed to generalize.

In contrast, radiometric features demonstrate physical generalization. Despite differing sensor wavelengths (1550nm vs 905nm), woody biomass exhibits higher intensity than broadleaf foliage in NIR/SWIR spectra. Standard scaling allows models to leverage this relative contrast without calibration.

Without intensity, models rely heavily on verticality (17.9%)—a geometric proxy for "trunk" (Fig. 6). When intensity is available, verticality reliance drops to 10.0% while intensity becomes dominant (18.9%). This suggests radiometric models identify wood by material properties, whereas geometric models identify wood by shape - a shape often occluded or distorted in aerial scans.

The upper canopy performance advantage (Table II) further illustrates this. Geometric descriptors require dense local neighborhoods to define coherent shapes; in sparse occlusion, these disintegrate. Intensity, being point-wise, remains robust even when branches yield only isolated returns.

Visual inspection reveals that geometry-only predictions exhibit fragmented wood structures (Fig. 3c-e), while radiometric augmentation maintains continuous connectivity (Fig. 3f). This structural coherence is critical for QSM reconstruction but is not captured by point-wise metrics like F1 score, highlighting a limitation in standard evaluation frameworks for this application.

## V. CONCLUSION

We evaluated leaf-wood segmentation under substantial domain shift - training on European TLS data (1550nm) and testing on North American RPA-LS data (905nm) - representing cross-site, cross-platform, and cross-species transfer. Our results demonstrate that geometry-only methods, including state-of-the-art deep learning architectures, fail to generalize when scan patterns and viewing geometries change.

We challenge the prevailing assumption that excluding radiometric features improves cross-sensor applicability. Radiometric features - particularly intensity - encode biological reflectance contrasts that persist across sensor wavelengths and viewing angles, providing more robust transfer than geometric descriptors. Incorporating these features improved F1 from 0.56 to 0.61 and more than doubled wood recall from 0.16 to 0.35.

Critically, we demonstrate that standard point-wise metrics (F1, Jaccard Index) can mask structural fragmentation that compromises downstream QSM reconstruction. Geometry-only methods achieve superficially acceptable F1 scores while producing disconnected branch segments, whereas radiometric augmentation preserves continuous tree architecture. This finding underscores the need for application-specific evaluation frameworks that assess structural coherence, not just classification accuracy.

Future work will focus on improving segmentation through enhanced data augmentation, hybrid architectures, and topology-preserving postprocessing methods. The ultimate goal is to integrate these methods into operational QSM-based biomass estimation pipelines, where improved leaf-wood separation translates to more accurate above-ground biomass estimation. Extension to coniferous species remains an important validation step for broader applicability.